# Artificial intelligence application in lymphoma diagnosis with Vision Transformer using weakly supervised training

Nghia (Andy) Nguyen[1], Amer Wahed[1], Andy Quesada[1], Yasir Ali[1], Hanadi El Achi[1], Y. Helen Zhang[1], Jocelyn Ursua[1], Alex Banerjee[1], Sahib Kalra[1], L. Jeffrey Medeiros[2], Jie Xu[2]

[1]Department of Pathology and Laboratory Medicine, The University of Texas McGovern Medical School, Houston, Texas

[2]Department of Hematopathology, The University of Texas MD Anderson Cancer, Houston, Texas

**Corresponding Author**

Nghia (Andy) Nguyen, MD

Department of Pathology and Laboratory Medicine

The University of Texas McGovern Medical School

6431 Fannin, MSB 2.224, Houston, TX, 77030, USA

Email: nghia.d.nguyen@uth.tmc.edu

**ABSTRACT**

Vision transformers (ViT) have been shown to allow for more flexible feature detection and can outperform convolutional neural network (CNN) when pre-trained on sufficient data. Due to their promising feature detection capabilities, we deployed ViTs for morphological classification of anaplastic large cell lymphoma (ALCL) versus classic Hodgkin lymphoma (cHL). We had previously designed a ViT model which was trained on a small dataset of 1,200 image patches in fully supervised training. That model achieved a diagnostic accuracy of 100% and an F1 score of 1.0 on the independent test set. Since fully supervised training is not a practical method due to lack of expertise resources in both the training and testing phases, we conducted a recent study on a modified approach to training data (weakly supervised training) and show that labeling training image patch automatically at the slide level of each whole-slide-image is a more practical solution for clinical use of Vision Transformer.  Our ViT model, trained on a larger dataset of 100,000 image patches, yields evaluation metrics with significant accuracy, F1 score, and area under the curve (AUC) at 91.85%, 0.92, and 0.98, respectively. These are respectable values that qualify this ViT model, with weakly supervised training, as a suitable tool for a deep learning module in clinical model development using automated image patch extraction.

**INTRODUCTION**

Machine learning consists of software that can learn from and make predictions on data - i.e., "gives software the ability to learn without being explicitly programmed." Numerous machine learning methods have been attempted in the past with varying degrees of success, including decision trees, cluster analysis, support vector machines, random forests, Bayesian analysis, regression analysis, neural networks, and large language models (LLMs) [1]. Deep learning (DL) is the most recent and most disruptive method of machine learning, based on neural networks and LLM. Currently, many large companies (e.g., Google, Microsoft, OpenAI) are analyzing large volumes of data for business analysis and decision-making, utilizing deep learning technology [2, 3]. The application of DL to digital pathology has shown potential that may have an impact on personalized diagnostics and treatment. Breakthroughs in DL occurred in 2006, enabling it to outperform other machine learning models. DL algorithms include two critical features: first, unsupervised learning enables a network to be fed with raw data (with no known outcomes) and to automatically discover the representations needed for detection or classification; and second, supervised learning extracts high-level and complex data representations. DL has benefited substantially from supporting hardware that facilitates parallel computation, a distinct feature of matrix multiplication, through a graphics processing unit (GPU) [4], which performs better than a central processing unit (CPU).

Lymphoma is a hematological malignancy, and the current classification of lymphoid malignancies by the World Health Organization is extensive [5]. Worldwide, 280,000 people have been diagnosed with lymphoid neoplasms annually. Lymphoma is typically first suspected based on its growth pattern and the cytologic features of the tumor cells, evaluated by light microscopy of hematoxylin-eosin-stained (H&E) tissue sections. Immunophenotypic features are typically needed for diagnosis, as determined by flow cytometry and/or immunohistochemical analyses. Additionally, cytogenetic results, molecular findings, and clinical features are often necessary to confirm the diagnosis or for prognostication. Anaplastic large cell lymphoma (ALCL) and classic Hodgkin lymphoma (CHL) are two distinct types of lymphoma that share some overlapping features (Fig. 1) [5]. ALCL is a rare subtype of peripheral T cell lymphoma

4(PTCL) that can present in different forms, including primary cutaneous ALCL and systemic ALCL. CHL is a more common type of lymphoma, characterized by the presence of Hodgkin-Reed-Sternberg cells in lymphoid tissue. Both types of lymphoma are marked by the presence of CD30 on the surface of the neoplastic cells, among other characteristic features. These neoplasms require careful diagnosis as different treatments are required to manage each disease effectively. The choice of treatment depends on the specific characteristics of the lymphoma and the patient's overall health. Due to subtle differences in histologic features between these two types of lymphomas, their diagnosis can represent a challenge to pathologists. Applying automated diagnostic tools that utilize whole-slide imaging (WSI) can help pathologists enhance diagnostic precision and efficiency. Earlier attempts to classify histologic images using features such as nuclear shape, size, and texture were largely unsuccessful [6], prompting the adoption of DL algorithms, particularly convolutional neural network (CNN) and vision transformer (ViT), which now represent the state-of-the-art in computer vision [7,8].

Convolution in CNN is an operation in image processing that uses filters to modify or detect specific characteristics of an image, such as smoothing, sharpening, intensification, or enhancement. Mathematically, a convolution is done by multiplying the pixel value in the image patch by a filter matrix (kernel) to yield a dot product. By moving the filter across an input image, one obtains the final output as a modified filtered image [9]. Recently, ViTs were shown to be capable of outperforming CNNs when pretrained on sufficient amounts of data [10-15]. In comparison to CNNs, ViTs have a weaker locality bias and therefore allow a more general feature detection (multi-modal data). ViTs are direct applications of transformer models based on a Large Language Model (such as Chat GPT, and Co Pilot) to vision analysis (Fig. 2). While the transformer architecture (based on Large Language Model) has become the de-facto standard for natural language processing tasks, its applications to computer vision remain limited. It was shown that applying Transformer algorithm directly to sequences of image patches can perform very well on image classification tasks. In 2021, Alexey Dosovitskiy et al. successfully deployed a ViT model for image analysis [16]. They demonstrated that a pure transformer applied directly to sequences of image patches can achieve high performance on image classification tasks. The pixel is the most basic unit of measurement in an image, but calculating every pixel relationship in a normal image would be memory intensive. Instead, ViT takes several steps, which require dividing the entire image into a grid of small image patches, then applying linear projection to embed each patch, with consideration for the position of each image patch in the image; each embedded patch becomes a token, and the resulting sequence of embedded patches is passed to the transformer encoder (Fig. 2). Self-attention in ViT allows each part of an image to relate (pay attention) to other parts of image, regardless of the distance between them. The input patches are processed using multi-head attention, and the output is fed into a multilayer perceptron (MLP), which produces the resultant classes, such as tumor types, in a tumor classifier. Due to their promising feature detection capabilities, we deployed ViTs for morphological classification of anaplastic large cell lymphoma (ALCL) versus classic Hodgkin lymphoma (cHL), using morphology data from whole-slide-images (scanned images from microscopic slides). In a previous study [17], we showed that ViT achieved a high diagnostic accuracy and F1 score on an independent test set. In that study, we used manually annotated image samples for training and testing (supervised learning), a time-consuming task that requires expert knowledge from pathology specialists. Our vision transformer model demonstrated accurate and consistent performance in classifying ALCL and cHL. On the independent test set, at Epoch 200, the model achieved a diagnostic accuracy of 100% and an F1 score of 1.0, matching the performance of our previously developed CNN model [18], which was also deployed on the same dataset. A confusion matrix was constructed to evaluate the model's final performance, showing no misclassifications. Possible reasons for the high accuracy of ViT in that study included: manual labelling of cases by hematopathologists, rigid inclusion constraints of WSI (all WSIs were scanned by the same scanner in the same laboratory), and only two types of lymphomas to predict by the model.

Traditional works on computational microscopic pathology are based mainly on supervised training algorithms [19]. These algorithms usually require pathologists' annotation of image patches on whole slide images (WSIs). The selected image patches are then used to train a supervised machine learning algorithm for inference tasks on histopathology images. Owing to the requirement of manual annotation by pathologists, such methods have limitations in real-world applications because accurate annotations from pathologists are difficult to obtain. This difficulty applies to both training and testing data. In light of the weaknesses of such methods, 'weakly-supervised training algorithms' has been developed in recent years [20]. Instead of requiring pathologists' detailed annotation at image patch-level, such algorithms rely on automated image extraction and require only one annotation per WSI (slide-level annotation), which can be easily obtained directly from the patients' electronic health records. In this study, we attempt to use weakly-supervised learning, in which all image patches in a case under consideration are automatically annotated with the given diagnosis of the case. This approach is expected to significantly decrease training time and does not require specialist expertise in annotation since the annotation process is automated. The results from this study would be the first example of using ViTs for lymphoma diagnosis with weakly-supervised training.

**MATERIALS AND METHODS**

We conducted a retrospective collection of cases with a previous diagnosis of cHL and ALCL, as defined by the current World Health Organization criteria [5], at our institutions from 2017 to 2025. We reviewed the morphological characteristics of each case. A total of 60 lymphoma cases (30 ALCL and 30 cHL) were collected from University of Texas, McGovern Medical School Laboratory and The University of Texas MD Anderson Cancer Center Hematopathology Laboratory. Whole-slide-image files were obtained by scanning the H&E glass slides using the Epredia P1000 scanner (3DHistech, Budapest, Hungary) with a scanning capacity for 1,000 glass slides, which produces high-quality images, full automation for focus, calibration, brightness and contrast settings, with tissue shape detection to outline and scan non-rectangular regions of interest for shorter turnaround times (Fig. 3). The total scan time for a 15 × 15 mm benchmark scan area at 40x resolution is less than or equal to 62 seconds. The images were acquired and converted to Slide Viewer Software (SVS) format (3DHistech, Budapest, Hungary) which was used to display the images. From each WSI, an average of 830 image patches with dimension 100 × 100 were extracted at 20x magnification, yielding a total of 50,000 image patches. Image patches were extracted from segmented areas of scans automatically using weakly-supervised training, in which all images in a case under consideration are automatically annotated with the given diagnosis of the case. Two automated image patch extraction methods were used (Fig. 4, Fig. 5). The first method is Python library FAST [21], an open-source framework developed by researchers at the Norwegian University of Science and Technology (NTNU), specifically useful for automated extraction of image patches utilizing multi-core CPUs and GPUs. The WSIs were segmented by FAST to obtain only the tissue area prior to image extraction. The second method is QuPath software [22], an open-source software developed at the University of Edinburgh, currently widely used for quantitative pathology and bioimage analysis. This software allows for automatic image extraction from annotated areas in WSIs, using Groovy script, an object-oriented language and is fully interoperable with Java [22].

Approximately 50,000 image patches (100x100 pixels) were extracted automatically in each method using weakly-supervised training, FAST Python library, and QuPath for setting up separate datasets. These patches were also combined to set up larger datasets of 100,000 image patches for analysis. The prediction performance of the ViT model using these two image patch extraction methods was also compared against each other to detect any discrepancy. The extracted patches were used for training, and testing with a ViT model previously developed by our research group. Of these, 90,000 (90%) were used for training, and 10,000 (10%) were used for testing. The cases were divided into two cohorts, with 30 cases (50,000 image patches) for each diagnostic category. For the ViT model, the training set included 45,000 image patches from ALCL cases and 45,000 from cHL cases. The test set consisted of 5,000 image patches from ALCL cases and 5,000 from cHL cases.

A 7-fold validation was used for testing [7], in which 10% image patches were used as test patches each time with the remaining 90% image patches used for training. The validation was performed seven times to rotate all image patches to testing category. This approach is used to ensure that the model does not predict a diagnosis based on memorizing the pattern from a particular case.

The ViT algorithm was written in Python using Torch, Torchvision, and Sklearn.metrics libraries (Python Software Foundation, Wilmington, Delaware, USA). Parallel processing was performed using dual NVIDIA RTX Pro 6000 Blackwell Max-Q GPU (NVIDIA Corp, Santa Clara, California, USA) with 96 GB of memory in each GPU for a total of 192 GB memory, and 24,064 CUDA (Compute Unified Device Architecture) cores in each GPU for a total of 48,128 cores. Software methods to fully utilize multiple GPUs include Pytorch libraries for DataParallel and Automatic Mixed Precision, Windows Subsystem for Linux 2 (WSL2) which allows for running a Linux environment directly on Windows for faster AI training time. The CPU is Intel Xeon W3-2525 CPU; 256 GB RAM. Windows 11 Professional for Workstations, 64-bit was the operating system (Microsoft Corp, Redmond, Washington, U.S.A.). The Institutional Review Board Ethics Committee approved the protocol for this retrospective study under the principles outlined in the Declaration of Helsinki. All cases in the study were deidentified to attain patient confidentiality, i.e., all H&E slides were scanned to obtain WSIs having consequential numbers without patient information. The WSI numbers were then associated with known diagnosis from previous reports, to be used as a master table for the study.

To assess prediction performance of the ViT model, binary classification metrics were used for evaluating the performance of our model which predicts one of two possible outcomes (ALCL or cHL). The metrics consist of accuracy, F1 score, and area under the curve (AUC). The equations for these quantities are as following [23]:

Accuracy= (TN + TP) /(TN + FN + FP + TP)

Where: TN=True Negative

TP=True Positive

FN=False Negative

FP=False Positive

F1 Score = 2 (Precision x Recall)/(Precision + Recall)

Where: Precision=TP/(FP +TP)

Recall=  TP/(FN + TP)

Area Under the Curve (AUC):

Where: AUC is the area under the Receiver Operating Characteristic (ROC) curve with a maximum value of 1.00. AUC helps us to understand how well the model separates the true-positive cases from the false-positive cases

Remoting to the DL workstation, where all datasets and codes reside, is used to let study group members get remote access to the workstation and run train/test image patches regardless of their locations which are spread over several clinical institutions. Quick Start guides and individual tutorials are given to study group members to be familiar with the remoting process. Results of classification metrics for various datasets are compiled from those obtained by group members.

**RESULTS**

Table 1 summarizes the results of this study, consisting of evaluation metrics (accuracy, F1 score, AUC) and training time for varying size of datasets from 25k to 100k. Overall, components of the evaluation metrics improve slightly from smaller datasets (25k) to intermediate ones (50k). Significant improvements are observed with progressive metrics for the largest datasets (100k).

They are the means obtained from 7-fold validation. The best evaluation metrics for datasets of 100k image patches showed accuracy, F1 score, AUC at 91.85%, 0.92, and 0.98, respectively. These are respectable values that qualify the ViT model, with weakly supervised training, as a DL tool suitable for clinical use. The evaluation metrics also showed that the two automated image extraction methods (FAST vs. QuPath) yielded image datasets with comparable prediction results. Also noted is the increase in training time for datasets with increasing sizes, as expected. Using our hardware/software setup for this study, the training time was maximum at approximately 5 hours for datasets of 100k image patches.

Production Module for Testing a New Case

A production protocol for testing unknown images was established for the ViT trained model to offer diagnosis prediction for new (unknown) images. Figure 6 illustrates a typical screen display for 5 unknown images patches. These image patches were obtained automatically from a WSI with unknown diagnosis (by any methods, including the two described in this article, FAST vs. QuPath). Five predicted diagnoses were shown by the pretrained ViT model, one for each image patch. The final diagnosis for the case was also given using majority voting, in which at least 3 of 5 predictions have to agree to be the final diagnosis [7]. In this example, all five image patches were predicted to be ALCL, and a final diagnosis of ALCL was suggested. Note that the run time for this production module is very short (1.45 sec in a typical run) since it utilized the pretrained model and did not have to go through training every time testing was needed.

**DISCUSSION**

To the best of our knowledge, our described research represents the first study using a ViT model for lymphoma diagnosis with detailed comparisons of the model's prediction in different training modes (fully supervised vs. weakly-supervised). Using weakly-supervised training (labelling at slide level), the evaluation metrics (accuracy, F1 score, and AUC) progressively improved with increasing the number of image patches used in training datasets: 25k, 50k, and 100k. Our ViT model, trained on the largest datasets of 100,000 image patches, yielded significant evaluation metrics with accuracy, F1 score, and AUC at 91.85%, 0.92, and 0.98, respectively. These are respectable values that qualify the ViT model, with weakly supervised training, as a suitable tool for DL development using automated image patch extraction.

Development of Vit models using fully supervised training is not a practical method due to lack of expertise resources in both training and testing phase. In this study, we used two automated methods to extract image patches and labeled them with diagnosis at the slide level, a technique used in weakly-supervised training. The two methods (FAST vs. QuPath) yielded image datasets with comparable prediction results for the ViT model.

While CNN remains the most widely adopted architecture for medical image classification, due in part to its mature tooling and efficiency on smaller datasets, ViT has shown growing promise in the field of computer vision. Notably, ViT lacks the strong inductive biases of CNN, such as local spatial hierarchies, instead relying on global self-attention mechanisms. This design allows ViT to capture long-range dependencies and potentially generalize better to multimodal or heterogeneous data. Its performance is often constrained by the algorithm's reliance on substantially larger datasets to achieve competitive performance, stemming from its data-hungry nature. Transformer-based models are also computationally intensive, particularly when processing large images, such as WSIs, where the self-attention mechanism incurs a quadratic computational cost that is directly proportional to the input size. These limitations have historically restricted the application of ViT in digital pathology, where images can span gigapixels. Nonetheless, our study uniquely shows that with proper data curation and task formulation, ViT algorithms can perform comparably to CNNs in distinguishing lymphoma types. These findings have proven to hold true even on moderately large datasets (100k) as seen in our current study, challenging the prevailing view that ViT is restricted to only very large-scale pretraining scenarios in which patches in image datasets have to be in the millions range.

Our current ViT prototype is a proof of concept for using DL as an assistance tool in clinical diagnosis. A busy pathologist may need a "second opinion" by DL for a difficult case in which a distinction between ALCL and cHL is not clear. The ViT model would also be useful in both on-going and "look-back" QA protocols. If a discrepancy is seen between the pathologist's diagnosis and the ViT model prediction, a second look by the pathologist would be warranted to assure that nothing has been overlooked in the diagnostic process. Conversely, there is no need for a second look if the pathologist's diagnosis and the ViT model prediction agree with each other.

The main limitations of the current study are: (a) inclusion of only two lymphoma types (ALCL and cHL); and (b) only image patches from WSIs of H&E slides. No immunohistochemical (IHC) slide results were considered for data input to the ViT model. Future directions will explore more lymphoma types and additional data from other modalities such as IHC analysis and molecular results. Additionally, performance should be validated across larger and more diverse datasets, including variations in stain types, scanner brands, and tissue heterogeneity. Cross-institutional validation will also be crucial to assessing model robustness in real-world settings. Other limitations of the current study include: (a) diagnosis of all image patches in a WSI is taken from reported diagnosis of the case, without further review to ensure that such diagnosis is correct; (b) no normalization of stains to reduce staining variability across histopathological slides; (c) no consideration for different magnifications to possibly enhance the classification metrics of the model (only 20x is used in the study). Future studies would require preview of the cases by a panel of hematopathologists to assure the accuracy of the reported diagnosis, normalization of stains [24], and using a multi-resolution model (such as 10x, 20x, 40x) to have image patches at various resolutions [24].

**CONCLUSION**

Our findings support a proof-of-concept for the feasibility and effectiveness of ViTs for lymphoma classification tasks in hematopathology. To the best of our knowledge, the results from this study are the first in using ViTs for lymphoma diagnosis with weakly-supervised learning. These results would significantly expand development of ViT models with more training data for lymphoma diagnosis. Weakly-supervised learning, as applied to ViT modelling, is expected to significantly decrease modelling time and does not require specialist expertise in annotation since the process is automated. Given the scalability, attention-based modeling, and adaptability to weakly-supervised training, the ViT-based architecture represents a promising avenue for future developments in computational pathology. Beyond predictive classification, ViT models hold immense potential as adjuncts in medical education, enabling trainees in pathology to engage in highly accurate computational tools and develop a more nuanced understanding of histologic features relevant to classification. As the field of pathology continues to adopt digital platforms, ViT-guided slide review may supplement both self-directed learning and traditional, instructor-led microscopy sessions, improving overall efficiency and proficiency. In clinical practice, ViT could provide utility as a diagnostic support tool, offering reassurance and reinforcement to pathologists in cases where a diagnosis is uncertain. Furthermore, by highlighting atypical or ambiguous morphologies, this model could serve to identify areas that may warrant additional review. This capability would be particularly valuable in settings with limited access to subspecialty expertise. As these applications materialize, integration into routine diagnostic workflows will require careful validation, optimization, and ongoing clinician oversight to ensure safe and effective use. Ultimately, with continued innovation in reducing their computational demands, ViT may emerge with an increasingly central role in the evolving landscape of digital diagnostics and clinical decision support.

**Author contributions**

Nghia Nguyen designed/conducted the study and wrote the manuscript. Yasir Ali, Hanadi El Achi, Helen Zhang, Jocelyn Ursua, Andy Quesada, Sahib Kalra, Alex Banerjee conducted the study and reviewed/edited the manuscript. Amer Wahed, L. Jeffrey Medeiros, and Jie Xu reviewed/edited the manuscript. All authors approved the manuscript in its final form.

**Conflict of Interest**

The authors declare that they have no conflicts of interest.

**REFERENCES**

1. Razzak MI, Naz S, Zaib A. Deep Learning for Medical Image Processing: Overview, Challenges and the Future. In: Dey N., Ashour A., Borra S. (eds) Classification in BioApps. First ed. Springer International Publishing; 2018:323-350

2. MIT Technol. Rev. 2013. Available at (last accessed on 10/30/18): https://www.technologyreview.com/s/513696/ deep-learning

3. LeCun, Y, Bengio, Y, Hinton, G. Deep learning. Nature. 2015;521:436–444

4. Janowczyk, A, Madabhushi, A. Deep learning for digital pathology image analysis: A comprehensive tutorial with selected use cases. J Pathol Inform. 2016;7:29.

5. WHO Classification of Tumours: Haematolymphoid Tumours, 5th Edition, Volume 11, 2024. WHO Classification of Tumours Editorial Board. 69008 Lyon, France: International Agency for Research on Cancer (IARC)

6. Choras RS. Feature Extraction for CBIR and Biometrics Applications. 7th WSEAS International Conference on Applied Computer Science. Vol. 7. 2007

7. Marsland S. Machine learning: an algorithmic perspective. Chapman and Hall/CRC, 2011.

8. Mitchell TM, Mitchell TM. Machine learning. Vol. 1. No. 9. New York: McGraw-Hill, 1997

9. Roy K, Banik D, Bhattacharjee D, Nasipuri M. Patch-based system for classification of breast histology images using deep learning. Computerized Medical Imaging and Graphics 71 (2019): 90-103.

10. Waqas, A, Bui, MM, Glassy, EF, et al. Revolutionizing Digital Pathology With the Power of Generative Artificial Intelligence and Foundation Models. Laboratory Investigation. Volume 103, Issue 11,100255, November 2023

11. Vaswani A, Shazeer, N, Parmar, N, et al. Attention Is All You Need. 31st Conference on Neural Information Processing Systems (NIPS 2017)

12. Vorontsov, E, Bozkurt, A, Casson, A, Shaikovski, G, Zelechowski, M, Severson, K, et al. A foundation model for clinical-grade computational pathology and rare cancers detection. Nature Medicine | Volume 30 | October 2024 | 2924–2935

13. Zhang, J, Lu, JD, Chen, B, et al. Vision transformer introduces a new vitality to the classification of renal pathology. BMC Nephrology (2024) 25:337

14. Khedr, OS, Wahed, ME, Al-Attar, AR, et al. The classification of the bladder cancer based on Vision Transformers (ViT). Nature Scientific Reports | (2023) 13:20639

15. Beyer, L, et al. Better plain ViT baselines for ImageNet-1k. Google Brain Research, Zurich https://github.com/google-research/big_vision.

16. Dosovitskiy, A, Beyer, L, Kolesnikov, A, et al. An image is worth 16x16 words: transformers for image recognition at scale. Proceedings of ICLR 2021

17. Rivera D, Banerjee A, Zhang R, El Achi H, Wahed A, Ho L, et al(2025) Vision Transformers for Diagnostic Classification of Lymphomas: A Matched Comparison with a Convolutional Neural Network, 21st Century Pathol, Volume 5 (1): 160

18. Rivera, D, Ali, K, Zhang, R, Mai, B, El Achi, H, Armstrong, J, Wahed, A, Nguyen, A. Deep Learning-Based Morphological Classification between Classical Hodgkin Lymphoma and Anaplastic Large Cell Lymphoma: A Proof of Concept and Literature Review, 21st Century Pathology, Volume 4 (1): 159

19. Mohri, M, Rostamizadeh, A,Talwalkar, A. Foundations of Machine Learning. MIT Press, Second Edition, 2018.

20. Li, Z, Cong, Y, Chen, X, et al. Vision transformer-based weakly supervised histopathological image analysis of primary brain tumors. iScience 26, 105872, January 20, 2023

21. Smistad, E, Østvik, A, Pedersen, A. High performance neural network inference, streaming, and visualization of medical images using FAST. IEEE Access, volume 7, 2019.

22. Bankhead, P., et al. QuPath: Open source software for digital pathology image analysis. Scientific Reports (2017). https://doi.org/10.1038/s41598-017-17204-5

23. Evaluation metrics in machine learning. https://www.geeksforgeeks.org/machine-learning/metrics-for-machine-learning-model/

24. Chaurasia, A, Toohey, PW, Harris, H, Hewitt, AW. Multi-resolution vision transformer model for histopathological skin cancer subtype classification using whole slide images. Computers in Biology and Medicine. Volume 196, Part A, September 2025, 110724.

**LEGENDS**

Figure 1. Representative histology of anaplastic large cell lymphoma (L) and classical Hodgkin lymphoma (R).

Figure 2. Vision Transformer (ViT) for image classification: the input image is split into smaller patches, where they are flattened, projected into vectors, and combined with positional embeddings. The Transformer Encoder utilizes N layers of multi-head attention, fully connected layers to process embeddings. A Multilayer Perceptron (MLP) is a feed-forward neural network where the information moves from the input layer to the output layer, passing through one or more hidden layers. The classifier outputs the final classification (e.g., "ALCL" or "cHL").

Figure 3. Lymphoma image processing workflow: 60 lymphoma cases stained with Hematoxylin and Eosin (H&E). Epredia P1000 scans slides at 40x magnification. Classical Hodgkin Lymphoma and Anaplastic Large T-Cell Lymphoma: two lymphoma types analyzed, shown with marked regions (blue and red squares). Image patches: 100x100 pixel patches were extracted from each lymphoma cohort for detailed analysis.

Figure 4. The workflow for two automated image extraction methods used in the study: FAST Python library (upper) and QuPath (lower).

Figure 5. Examples of automated image patch extractions: from classical Hodgkin Lymphoma cases (L) and from Anaplastic Large Cell Lymphoma (R).

Figure 6. Screen display of the production module, predicting diagnosis for 5 unknown image patches from the same WSI, also with final diagnosis using majority voting.

**TABLES AND FIGURES**

Table 1: Evaluation metrics and training time

| Dataset size | Accuracy | AUC | F1 Score | Training time |
|---|---|---|---|---|
| 25k (FAST) | 84.84% | 0.91 | 0.81 | 1 hr 3 min |
| 25k (QuPath) | 82.80% | 0.87 | 0.72 | 1 hr 13 min |
| 50k (FAST) | 84.48% | 0.96 | 0.89 | 2 hr 5 min |
| 50k (QuPath) | 84.65% | 0.98 | 0.87 | 2 hr 10 min |
| 100k (FAST+ QuPath) | 91.85% | 0.98 | 0.92 | 5 hr 8 min |

All results are means of measurements with 7-fold validation

**Fig. 1**

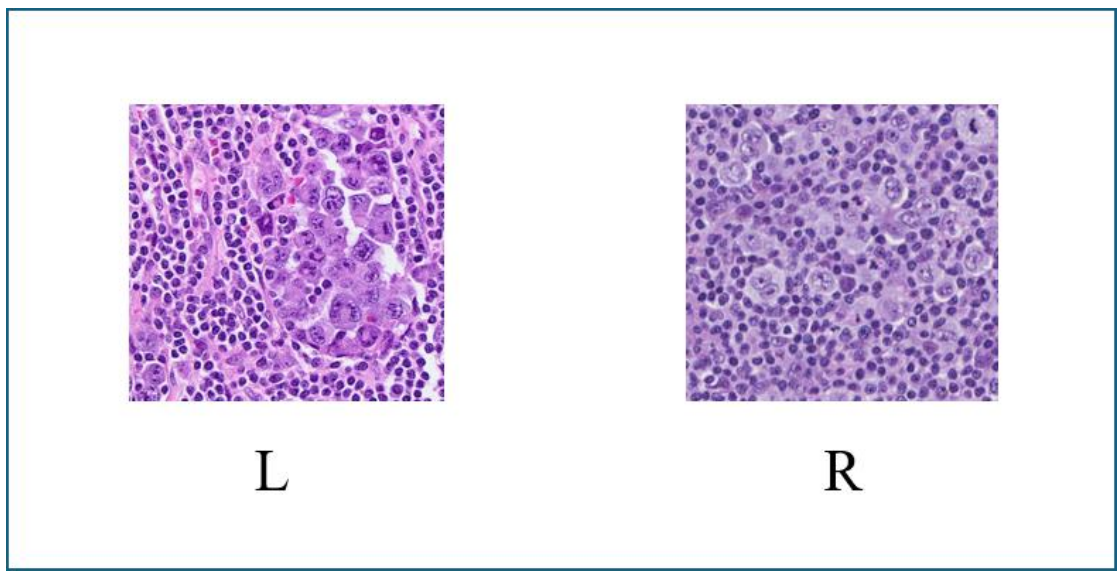

**Fig. 2**

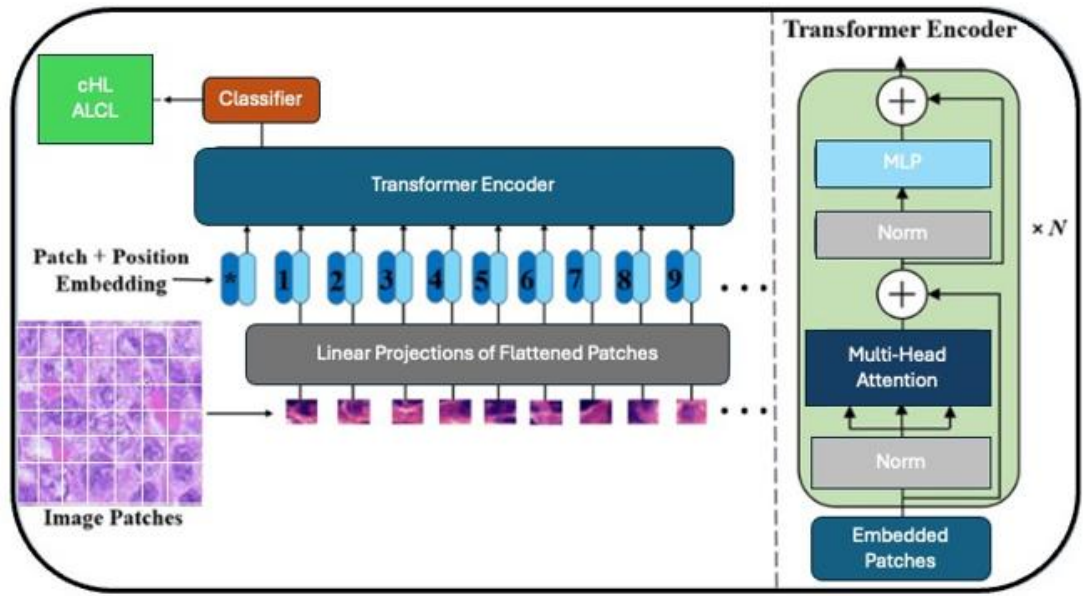

**Fig. 3**

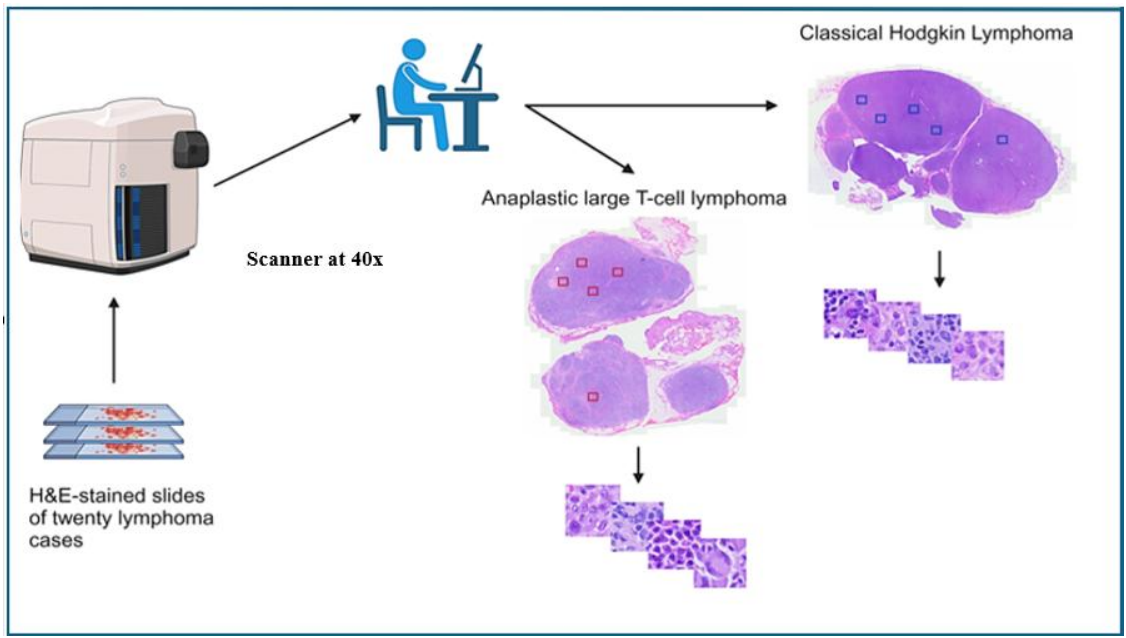

**Fig. 4**

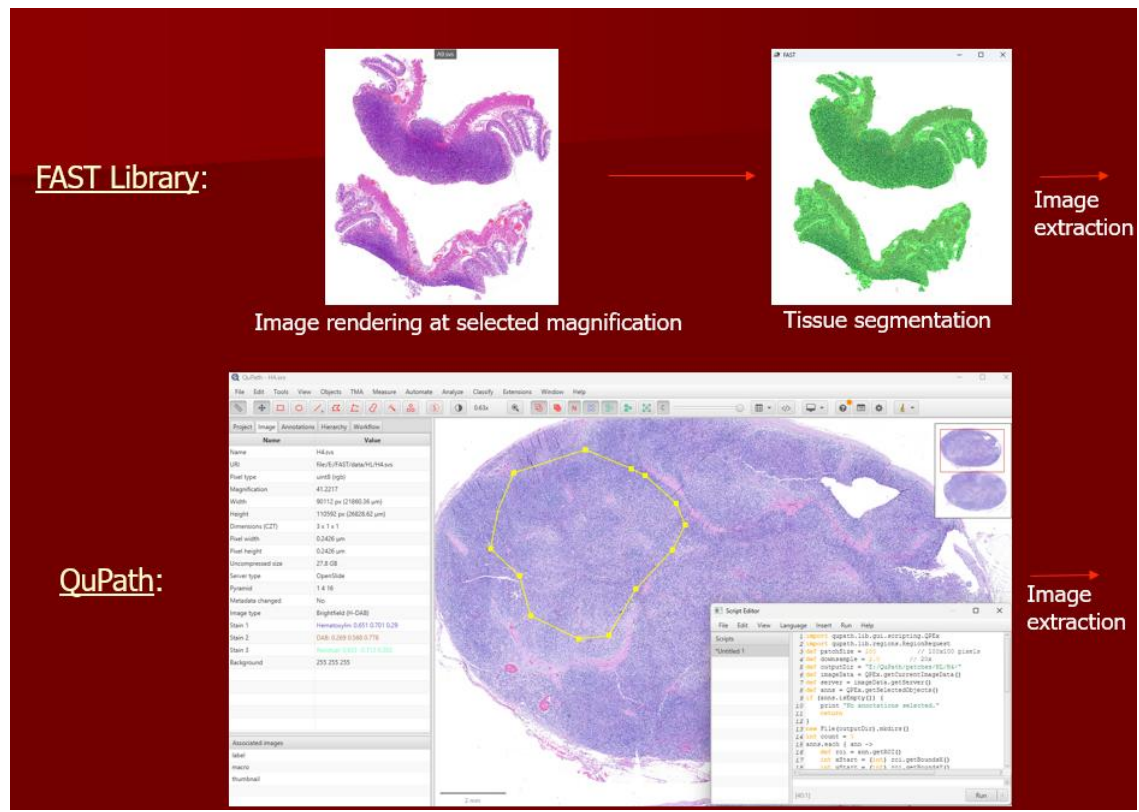

**Fig. 5**

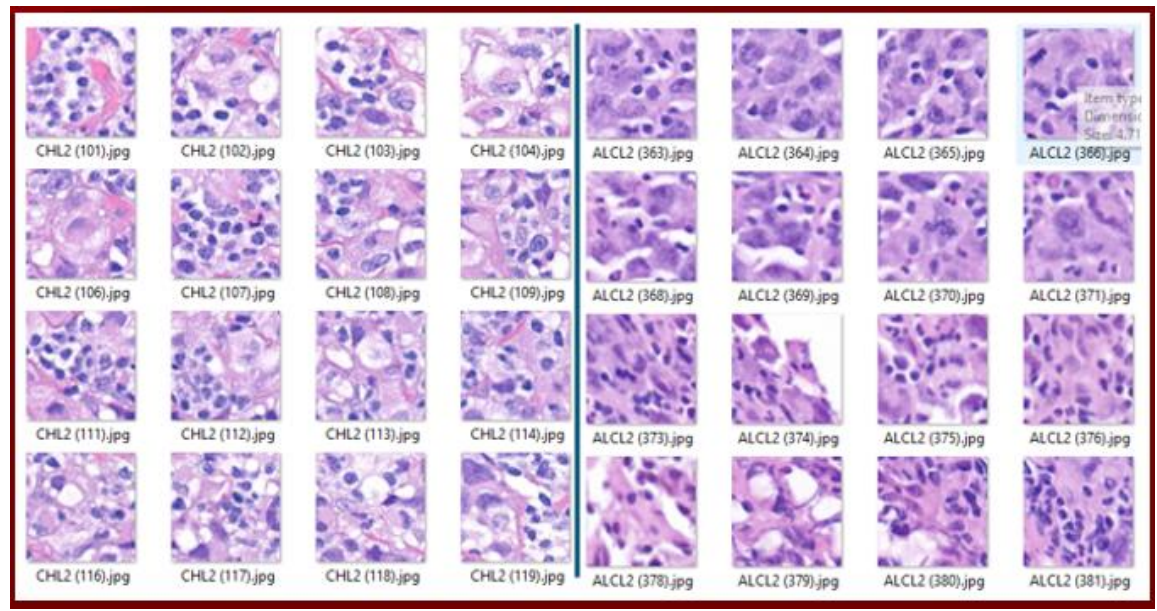

**Fig. 6**

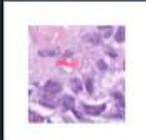